\documentclass[a4paper,11pt]{article}
\usepackage[utf8]{inputenc}

\usepackage{geometry}
\usepackage{setspace}
\usepackage{authblk}

\usepackage{amsmath}
\usepackage{amsfonts}
\usepackage{graphicx}
\usepackage{float}
\usepackage{subcaption}

\usepackage[numbers,sort,super]{natbib}

\usepackage{multicol}
\usepackage{multirow}
\usepackage{xcolor}
\usepackage{booktabs}

\usepackage[colorlinks=true, allcolors=blue]{hyperref}

\begin{document}

\title{\textbf{Geometric Graphs from Data to Aid Classification Tasks with Graph Convolutional Networks}}
\author[1]{Yifan Qian}
\author[2,3]{Paul Expert}
\author[1,*]{Pietro Panzarasa}
\author[4,5,*]{Mauricio Barahona}
\affil[1]{School of Business and Management, Queen Mary University of London, London, UK}
\affil[2]{Global Digital Health Unit, School of Public Health, Imperial College London, London, UK}
\affil[3]{World Research Hub Initiative, Tokyo Institute of Technology, Tokyo, Japan}
\affil[4]{Department of Mathematics, Imperial College London, London, UK}
\affil[5]{Lead Contact}
\affil[*]{Correspondence: m.barahona@imperial.ac.uk (M.B.), p.panzarasa@qmul.ac.uk (P.P.)}
\date{}

\maketitle

\newpage

\section*{Summary}
Traditional classification tasks learn to assign samples to given classes based solely on sample features. This paradigm is evolving to include other sources of information, such as known relations between samples. Here we show that, even if additional relational information is not available in the data set, one can improve classification by constructing geometric graphs from the features themselves, and using them within a Graph Convolutional Network. The improvement in classification accuracy is maximized by graphs that capture sample similarity with relatively low edge density. We show that such feature-derived graphs increase the alignment of the data to the ground truth while improving class separation. We also demonstrate that the graphs can be made more efficient using spectral sparsification, which reduces the number of edges while still improving classification performance. 
We illustrate our findings using synthetic and real-world data sets from various scientific domains.

\section*{Keywords}
classification tasks, machine learning, graph convolutional networks, graph neural networks, geometric deep learning, graph construction, graph sparsification, graph theory, network science, data science
\newpage

\section*{Introduction}
Classifying samples into a given set of classes is one of the fundamental tasks of data analytics~\cite{lecun2015deep}. In supervised machine learning, traditional methods train a classifier using a data set in which both features and class labels are observed for each sample. Once a classifier has been learned from the training data set, its parameters are optimized over a validation set. Then the model can be used to predict the class of unseen samples based on their features. Intuitively, a good classifier learns a representation of the data where samples belonging to different classes are well separated. 

In some instances, data sets contain additional information in the form of observed relational links between samples. For example, in a data set of scientific articles, each article will be described by features that encode its text, but we might also have information on citations between articles; in a data set of patients, each person will be associated with a series of clinical or socio-economic features, but we might also have information about their social interactions. Such relational information could be used in conjunction with the sample features to achieve the best possible class separation, and hence improved classification.
Graphs are a natural way to represent such relational links: the samples are viewed as the nodes of the graph, and the relationships between samples are formalized as edges. A large number of machine learning methods have been proposed to leverage the information in such graph structure. Graph Neural Networks (GNNs) is a nascent class of methods, which refers to a broad set of techniques attempting to extend deep neural models to graph-structured data~\cite{bronstein2017geometric}. GNN has witnessed success in a variety of research domains including computer vision~\cite{xu2017scene,landrieu2018large}, natural language processing~\cite{kipf2017semi,hamilton2017inductive,velickovic2018graph,peach2020semi}, traffic~\cite{li2018diffusion,yu2018spatio}, recommendation systems~\cite{ying2018graph,monti2017geometric}, chemistry~\cite{duvenaud2015convolutional,gainza2020deciphering} and many other areas~\cite{allamanis2018learning,qiu2018deepinf,zugner2018adversarial,choi2017gram,choi2018mime,li2018combinatorial}. For an in-depth review of GNNs, see Ref.~\cite{wu2020comprehensive}. 

Recently, work with Graph Convolutional Networks (GCNs)~\cite{kipf2017semi} has suggested that using a graph of samples in conjunction with sample features can improve classification performance when compared with traditional methods that use only features. Computationally, the graph allows the definition of a convolution operation that exchanges and aggregates the features of samples that are connected on the graph. If the graph and the features align well with the underlying class structure~\cite{qian2021quantifying}, the graph convolution operation homogenizes features of neighboring nodes, which will also tend to be more similar, while making less similar samples, which will be more distant on the graph, belong to other classes.  

In many instances, extra relational information in the form of a graph is not easily available. However, the intuition that nodes that are close in feature space tend to belong to the same class can still be exploited by constructing geometric graphs directly from the data features, and in doing so creating neighborhoods of similar samples. Such feature-derived graphs can then be used to aid and potentially sharpen the classification.  

\begin{figure}[H]
\centering
    \begin{subfigure}[c]{0.7\textwidth}
    \centering
    \caption{}
    \includegraphics[width=\textwidth]{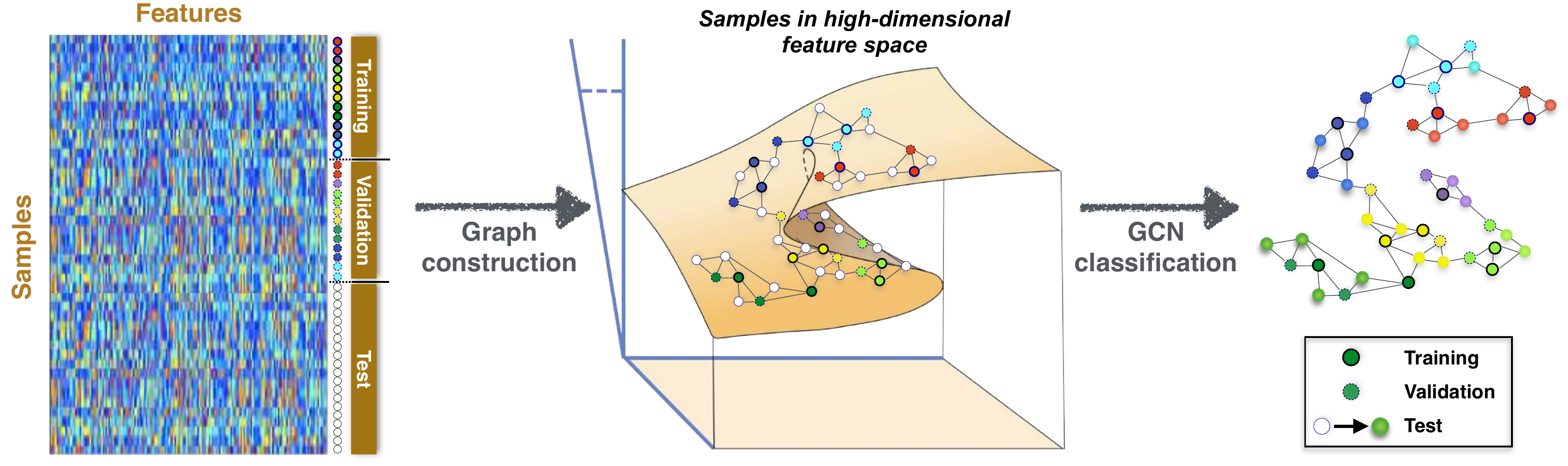}
    \label{fig:schematic_diagram}
    \end{subfigure}
    \begin{subfigure}[c]{0.55\textwidth}
    \centering
    \caption{}
    \centering
    \includegraphics[width=.9\textwidth]{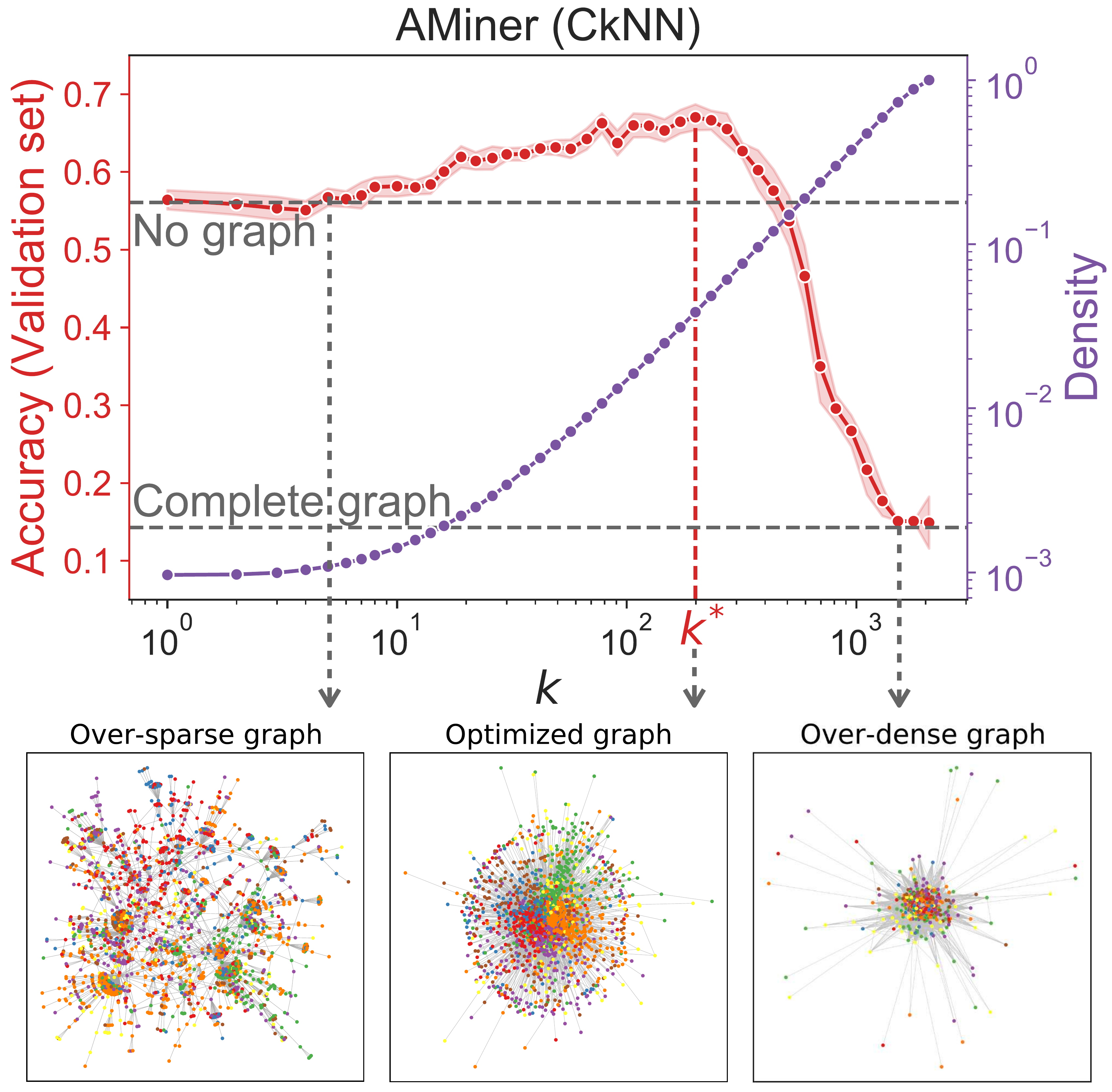}
    \label{fig:aminer_CkNN_densification}
    \end{subfigure}
    \caption{
    \textbf{Geometric graphs constructed from data features can aid sample classification.}\\
    (a) Workflow for GCN classification using feature-derived graphs. 
    (b) The validation set is used to search for graphs with optimized edge density---here illustrated with the AMiner data set and CkNN graph construction. In red, the GCN classification accuracy on the validation set as a function of the density parameter, $k$. The results are averaged over $10$ runs with random weight initializations; shaded region represents standard deviation. 
    As we sweep $k$ from `no graph' (MLP) to complete graph (mean field, random assignment), the classification accuracy on the validation set exhibits a maximum for a CkNN graph with density parameter $k^*$.  In purple, edge density of the CkNN graphs as $k$ is varied. Figures for all graph constructions and data sets are provided in Figure~S1.
    Also shown below, graph visualizations using the spring layout for over-sparse, optimized and over-dense graphs, with nodes colored according to their ground truth class.
    }
\end{figure}

Here, we explore the benefit of constructing geometric graphs from the features of the samples and using them within a GCN for sample classification (Figure~\ref{fig:schematic_diagram}). Graph construction, or inference, is a problem encountered in many fields~\cite{newman2018network}, from neuroimaging to genetics, and can be based on many different types of heuristics, from simple thresholding~\cite{zalesky2012use} or statistically significant group-level thresholding~\cite{lord2012functional} to sophisticated regularization schemes~\cite{omranian2016gene}. In general, the goal is to obtain graphs that concisely preserve key properties of the original data set as sparsely as possible, i.e., with a low density of edges.
In this work, we use several popular geometric graph constructions to extract graphs from data, and study how the classification performance depends on the graph construction method and the edge density. We find that there is a range of relatively low edge densities over which the constructed graphs improve the classification performance. Among the construction methods, we show that the recently proposed Continuous $k$-Nearest Neighbor (CkNN)~\cite{berry2019consistent} performs best for GCN classification. 
To gain further intuition about the role played by the graph in improving classification, we compute two simple measures: (i) the alignment of the convolution of graph and features with the ground truth; and (ii) the ratio of class separation in the output activations of the GCN. We show that the optimized geometric graphs increase the alignment and the class separation. Finally, we show that the graphs can be made more efficient using spectral graph sparsification~\cite{spielman2011graph}, which reduces the edge density of the optimized CkNN graphs while improving further the classification performance. 

\begin{table}[H]
\centering
\caption{
Classification accuracy (in percent) on the test set (averaged over $10$ runs with random initializations) for $7$ data sets with $8$ classifiers (four graph-less methods; GCN with four graph constructions). The standard deviation is reported in Table~S2. The top two results for each data set are bold. Overall, GCN with CkNN graphs displays the best performance. The density parameters of optimized graphs are reported in Table~S3.}
\label{table:results_classification}
\resizebox{.9\textwidth}{!}{
    \begin{tabular}{c|ccccccc|c}
        \toprule
        \textbf{Classifier} & \textbf{Constructive} & \textbf{Cora} & \textbf{AMiner} & \textbf{Digits} & \textbf{FMA} & \textbf{Cell}  & \textbf{Segmentation} &
        \textbf{Average improvement} \\
        \midrule
        MLP = GCN (No graph) &  42.1 &  54.2 &  54.4& 82.0 & 34.3 & 79.5&72.0  & --\\
        kNNC  &  31.4 & 38.2 & 28.0 & 88.3 & 30.6 & 58.7& 68.8 & $(-10.6)$ \\
        SVM &  40.0&55.9  &51.4  &87.7  &35.3  &81.5 &\textbf{87.7} & $(+3.0)$ \\
        RF & 36.3 & 56.1 &47.7  &83.0  &33.0  &\textbf{88.0}  &\textbf{88.8} & $(+2.1)$  \\
        \midrule
        GCN (kNN) & \textbf{53.9} & \textbf{66.4} & 59.2   & 92.0 &\textbf{35.6}  &83.8 & 83.5 & \textbf{$(+8.0)$} \\ 
        GCN (MkNN) & 45.2 & 64.1 &\textbf{61.8}  & \textbf{93.2} & \textbf{35.6}  & 84.0 & 83.0 & $(+6.9)$\\ 
        GCN (CkNN) & \textbf{51.1} & \textbf{66.6} & \textbf{61.6}  & \textbf{93.4} &  \textbf{36.0} & 84.0 & 83.9 & \textbf{$(+8.3)$}\\
        GCN (RMST) & 45.9 & 64.8 & 61.5 &  89.3 & 35.4 & \textbf{84.9}  & 83.0 & $(+6.6)$\\
        \bottomrule
\end{tabular}
}
\end{table}

\section*{Results}
\subsection*{Geometric Graphs Constructed from Data Features Can Aid Sample Classification}
We consider geometric graph constructions that fall broadly in two groups: (i) three methods based on local neighborhoods, i.e., $k$-Nearest Neighbor (kNN), Mutual $k$-Nearest Neighbor (MkNN) and CkNN~\cite{berry2019consistent} graphs; and (ii) a method that balances local and global distances measured on the Minimum Spanning Tree (MST), i.e., the Relaxed Minimum Spanning Tree (RMST)~\cite{beguerisse2013finding}. In all cases, we start from an MST to guarantee the resulting graph comprises a single connected component, and we add edges based on the corresponding distance heuristics. 
In each construction, a parameter regulates the edge density of the graph: $k$ in kNN, MkNN and CkNN, and $\gamma$ in RMST (see Methods for a full description of the methods).

For each data set and each graph construction, we find the edge density that maximizes the average GCN classification accuracy on the validation set by sweeping over $50$ values of the edge density, from sparse to dense. For each value of the density, we run the GCN classifier $10$ times starting from random weight initializations to compute the average accuracy.
Note that the two limiting cases are well characterized: the `no graph' limit corresponds to the Multilayer Perceptron (MLP); the `complete graph' limit is equivalent to mean field and leads to random class assignment~\cite{qian2021quantifying}.
Figure~\ref{fig:aminer_CkNN_densification} shows the classification performance of a GCN with a CkNN graph of increasing density applied to a data set of computer science papers (AMiner), which we use as our running example throughout. 
We find that adding relatively sparse graphs improves the classification accuracy, reaching a maximum increase of $10.9\%$ at an edge density of $0.039$ ($k^*=199$) on the validation set.
Once the edge density parameter is optimized on the validation set, we apply the GCN classifier to the test set and the test accuracy is recorded. In this case, the GCN yields an improvement of $7.2\%$ in classification accuracy on the test set with respect to MLP, as reported in Table~\ref{table:results_classification}.

We have investigated six real-world data sets from different domains, ranging from text (AMiner~\cite{qian2017citation,tang2008arnetminer}, Cora~\cite{sen2008collective}) to music track features 
(FMA~\cite{franceschi2019learning,defferrard2017fma}) to single-cell transcriptomics (Cell~\cite{velmeshev2019single}) to imaging (Digits~\cite{pedregosa2011scikit}, Segmentation~\cite{Dua:2019}). 
We have also studied one constructive data set with a well-defined ground truth based on a stochastic block model. For a detailed description of the data sets, see Supplemental Information 1 and Table~S1). 
We have compared the performance of four graph-less, feature-based classifiers (MLP, kNN classification (kNNC), Support Vector Machine (SVM) and Random Forest (RF)) to GCN classifiers with optimized feature-derived geometric graphs (Table~\ref{table:results_classification}). Our numerical experiments indicate that the GCNs with feature-derived graphs generally achieve better classification performance than graph-less classifiers. In particular, the CkNN graph construction achieves the highest accuracy improvement (+$8.3\%$ on average above MLP) across our seven data sets.

\begin{figure}[H]
\centering
    \begin{subfigure}[c]{0.45\textwidth}
    \centering
    \caption{}
    \includegraphics[width=\textwidth]{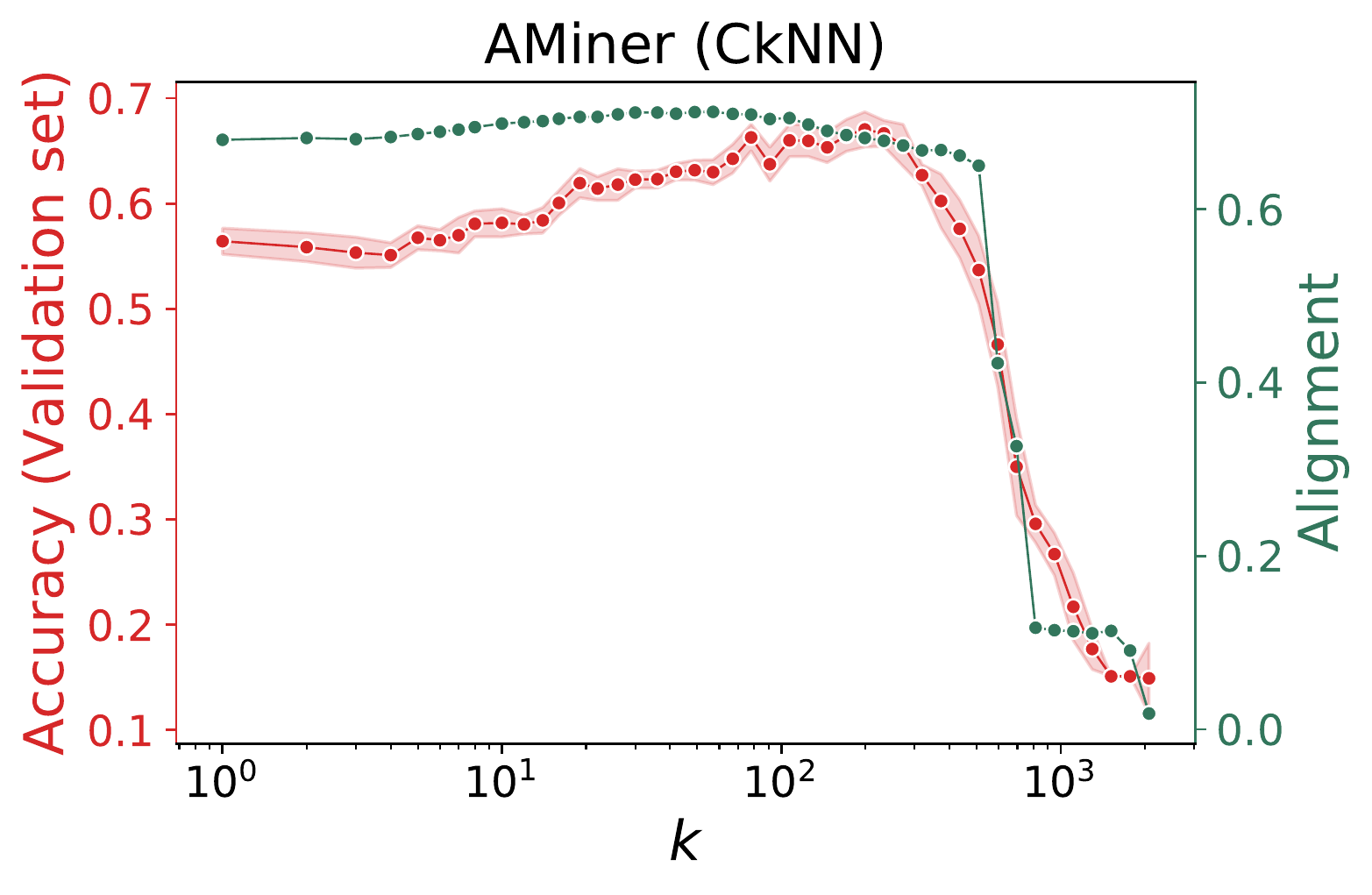}
    \label{fig:aminer_alignment}
    \end{subfigure} \\
    \begin{subfigure}[c]{0.45\textwidth}
    \centering
    \caption{}
    \includegraphics[width=\textwidth]{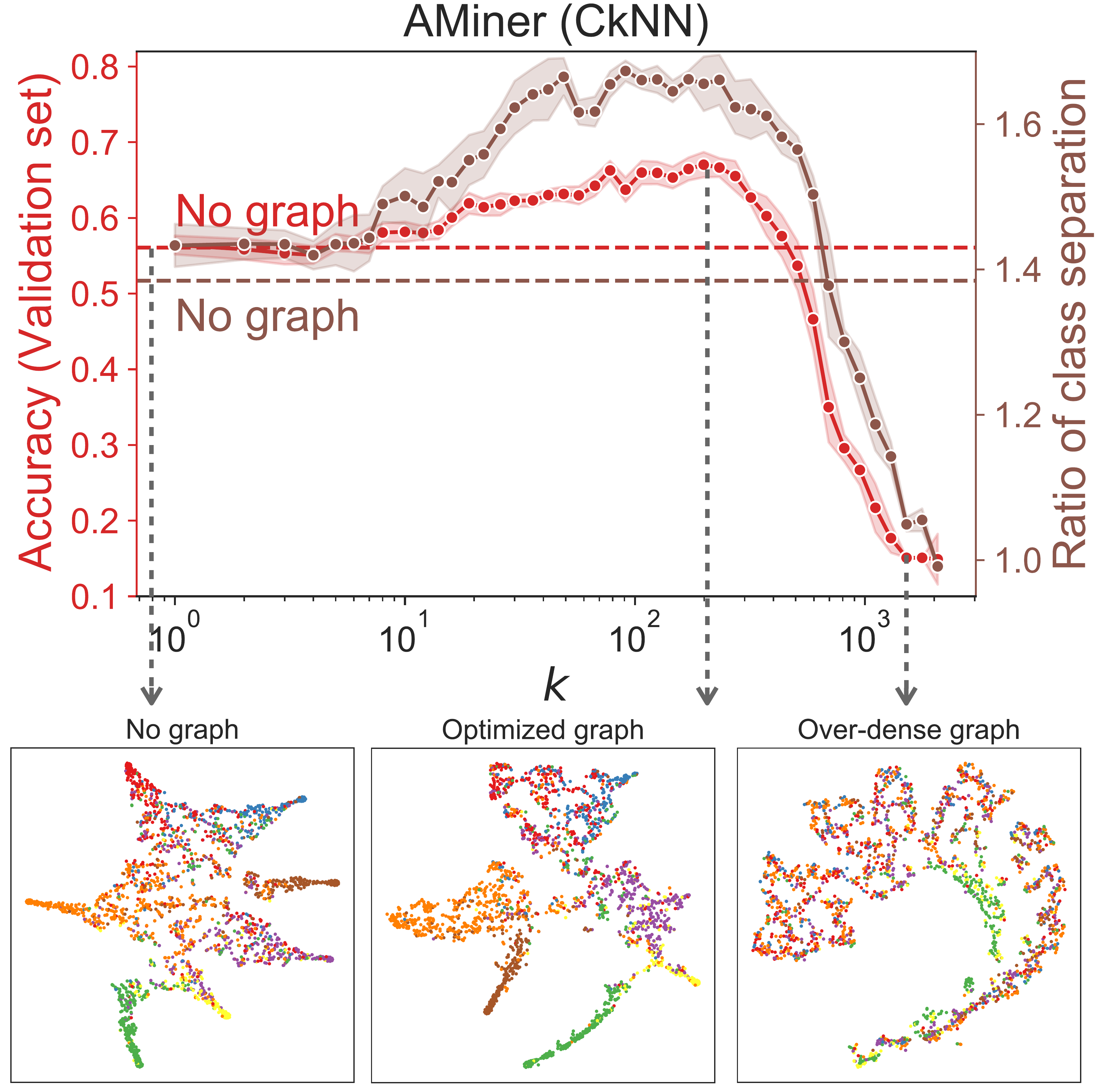}
    \label{fig:aminer_rcs}
    \end{subfigure}
    \caption{
    \textbf{The role of feature-derived graphs in classification.}\\
    (a)
    In green, we show the alignment (Equation~\eqref{eq:alignment}) of CkNN graphs for the AMiner data set as a function of the density parameter $k$.
    In red, classification accuracy as in Figure~\ref{fig:aminer_CkNN_densification}.
    The drop in classification accuracy corresponds to the drop in the subspace alignment. 
    Results for all graph constructions and data sets are given in Figure~S2. 
    (b)
    Ratio of class separation (Equation~\eqref{eq:rcs}) computed from the output activations of the GCN with CkNN graphs for AMiner data set as a function of the density parameter $k$, in brown. The results are averaged over $10$ runs with random weight initializations; shaded region is the standard deviation. The brown dashed line represents the RCS for the MLP, i.e., GCN with no graph. In red, classification accuracy, as in Figure~\ref{fig:aminer_CkNN_densification}.
    Below, we show two-dimensional t-SNE projections of the output activations of GCNs with no graph, optimized graph and over-dense graph. The nodes are colored according to the ground truth class labels. The optimized graph induces higher class separability, as shown by an increased RCS and better resolved t-SNE projection. 
    Results for all graph constructions and data sets are provided in Figure~S3.
    }
    \label{}
\end{figure}

\subsection*{The Role of Feature-derived Graphs in Classification}   
Our results show improved classification performance of GCNs with feature-derived geometric graphs of appropriate edge density. Indeed, over-sparse graphs perform close to MLPs, the `no graph' limiting case, whereas over-dense graphs are detrimental, as they approach the `mean field' limit that behaves like random class assignment. Hence there is a sweet spot of relatively low edge density where graphs improve the performance maximally. To gather further insight into the role of the constructed graphs in classification, we explore their properties from two complementary perspectives. 

\subsubsection*{Over-dense Graphs Degrade the Alignment of Graph-convolved Features with the Ground Truth} 
Consider the classification of $N$ samples with $F$ features into $C$ classes making use of a graph with adjacency matrix $A$.
In Ref.~\cite{qian2021quantifying} it was shown that good GCN performance requires a certain degree of alignment between the linear subspaces associated with the matrix of features, $X \in R^{N \times F}$, the adjacency matrix of the graph with self-loops, $\widehat{A} \in R^{N \times N}$, and the ground truth membership matrix, $Y\in R^{N \times C}$ (see Methods for a full description of GCNs).
Inspired by Ref.~\cite{qian2021quantifying}, we evaluate the alignment between the ground truth $Y$ and the graph-convolved features $X_A := \widehat{A}X$ as:
\begin{equation}
     S(X,\widehat{A},Y) = \cos (\theta_{1}(\mathcal{X_{A}}, \mathcal{Y}) ).
     \label{eq:alignment}
\end{equation}
Here $\theta_{1}(\mathcal{X_{A}}, \mathcal{Y})$ is the \textit{minimal principal angle}~\cite{bjorck1973numerical,golub2013matrix,knyazev2002principal}
between the column spaces of the matrices $\text{PCA}(X_A, p^*)$
and $\text{PCA}(Y, p^*)$, which contain the top Principal Components, as determined by the parameter $p^*$, of $\widehat{A}X$ and $Y$, respectively.  
The parameter $p^*$ is the ratio of explained variance that maximizes the Pearson correlation between the alignment~(Equation~\eqref{eq:alignment}) and the classification accuracy on the validation set. 

Figure~\ref{fig:aminer_alignment} shows the alignment~(Equation~\eqref{eq:alignment}) between the ground truth and the graph-convolved data for CkNN graphs of increasing density on the AMiner data set. We find that the reduction in classification accuracy induced by over-dense graphs is linked to a strong disruption of the subspace alignment $S\left(X,\widehat{A},Y \right)$. In the limit of the complete graph, the alignment approaches the value of $0$, i.e., the minimal angle $\theta_1 = \pi/2$ indicating that the two subspaces are orthogonal. Sparse graphs, on the other hand, induce a slight increase of the subspace alignment at the same time as improving the classification accuracy. The alignment and classification accuracy show good correlation for the AMiner data set: the Pearson correlation between alignment and accuracy (validation set) is $0.970$, obtained for a value of $p^*=0.4$. The same procedure has been carried out for all seven data sets, and the results are presented in the Figure~S2. The Pearson correlation coefficient between alignment and accuracy (validation set) ranges from $0.602$ (Segmentation) to $0.970$ (AMiner) with an average of $0.852$ over all $7$ data sets, thus indicating a good correspondence between the classification accuracy and the graph-induced alignment of data and ground truth.

\subsubsection*{Graphs with Optimized Density Increase the Ratio of Class Separation}
Another way of assessing the effect of the constructed graphs on classification is to study the inherent separability of the probabilistic GCN assignment matrix, i.e., the row-stochastic matrix $Z \in R^{N \times C}$ of output activations in Equation~\eqref{eq:two_layers_rule}. The effect of the graph on $Z$ reflects the quality of the classifier: a good graph should enhance the separation of samples from different classes while clustering together samples from the same class in $C$-dimensional space. 
We quantify the separability of the GCN mapping using $Z' \in R^{N \times 2}$, the
two-dimensional t-SNE~\cite{maaten2008visualizing} embedding of $Z$, on which we compute the ratio between the average inter-class and intra-class distances, denoted ratio of class separation (RCS):
\begin{equation}
    \text{RCS} = 
    \dfrac{ (\mathbf{1}^T ( D^{(Z')}\circ M^\text{inter} ) \mathbf{1} )/ (\mathbf{1}^T  M^\text{inter} \mathbf{1})} {(\mathbf{1}^T \left( D^{(Z')}\circ M^\text{intra}\right) \mathbf{1}) / (\mathbf{1}^T  M^\text{intra} \mathbf{1}) }.
    \label{eq:rcs}  
\end{equation}
Here, $D^{(Z')}$ is the Euclidean distance matrix for the t-SNE embedding $Z'$, i.e., $D^{(Z')}_{ij}=\left\|Z'_{i}-Z'_{j}\right\|_{2}$; the notation $\circ$ represents the Hadamard, element-wise, matrix product; $M^\text{inter}\in R^{N \times N}$ is the inter-class indicator matrix, i.e., $M^\text{inter}_{ij}=1$ if samples $i$ and $j$ belong to different classes and $M^\text{inter}_{ij}=0$ otherwise; and, conversely, $M^\text{intra}\in R^{N \times N}$ is the intra-class indicator matrix. Compactly, we have
\begin{align*}
M^\text{inter} &= \mathbf{1 1}^{T}-YY^{T}\\
M^\text{intra} &= YY^{T}-I_{N},
\end{align*}
where $I_{N} \in R^{N \times N}$ is the identity matrix and $\mathbf{1}$ is the $N$-dimensional vector of ones. 

Figure~\ref{fig:aminer_rcs} shows the RCS~(Equation~\eqref{eq:rcs}) computed from the output activation of GCNs with CkNN graphs of increasing density (AMiner data set). We observe a high correlation between RCS and the classification accuracy (validation set): the Pearson correlation coefficient for AMiner is $0.953$.
Similar figures for all data sets are shown in the Figure~S3. The Pearson correlation coefficient between RCS and accuracy (validation set) is high for all data sets, ranging from $0.876$ (Segmentation) to $0.976$ (Cora), with an average Pearson correlation coefficient of $0.938$ across all $7$ data sets.
These results indicate that sparse graphs unfold the data and facilitate class separation, as illustrated by the t-SNE plots and the increased RCS; on the other hand, over-dense graphs reduce separability and eventually converge to the mean field limiting value of RCS $=1$, i.e., when there is no distinction between inter- and intra-class separation.

\subsection*{Spectral Sparsification of Optimized Geometric Graphs Can Further Improve Classification}

Sparse graphs are generally favored over dense graphs, in particular for large data sets, as they are more efficient for both numerical computation and data storage. 
We investigate whether it is possible to sparsify the optimized geometric graphs obtained above, while preserving, or even improving, GCN classification performance.
Motivated by the key importance of spectral properties in graph partitioning~\cite{delvenne2010stability,lambiotte2014random}, we apply the Spielman-Srivastava sparsification algorithm (SSSA)~\cite{spielman2011graph} to our optimized CkNN graphs. The SSSA reduces the number of edges of a graph while preserving the spectral content of the graph Laplacian given by Equation~\eqref{eq:sparsification} (see Methods for a full description of the method).

\begin{figure}[H]
    \centering
    \begin{subfigure}[c]{0.45\textwidth}
    \centering
    \caption{}
    \includegraphics[width=\textwidth]{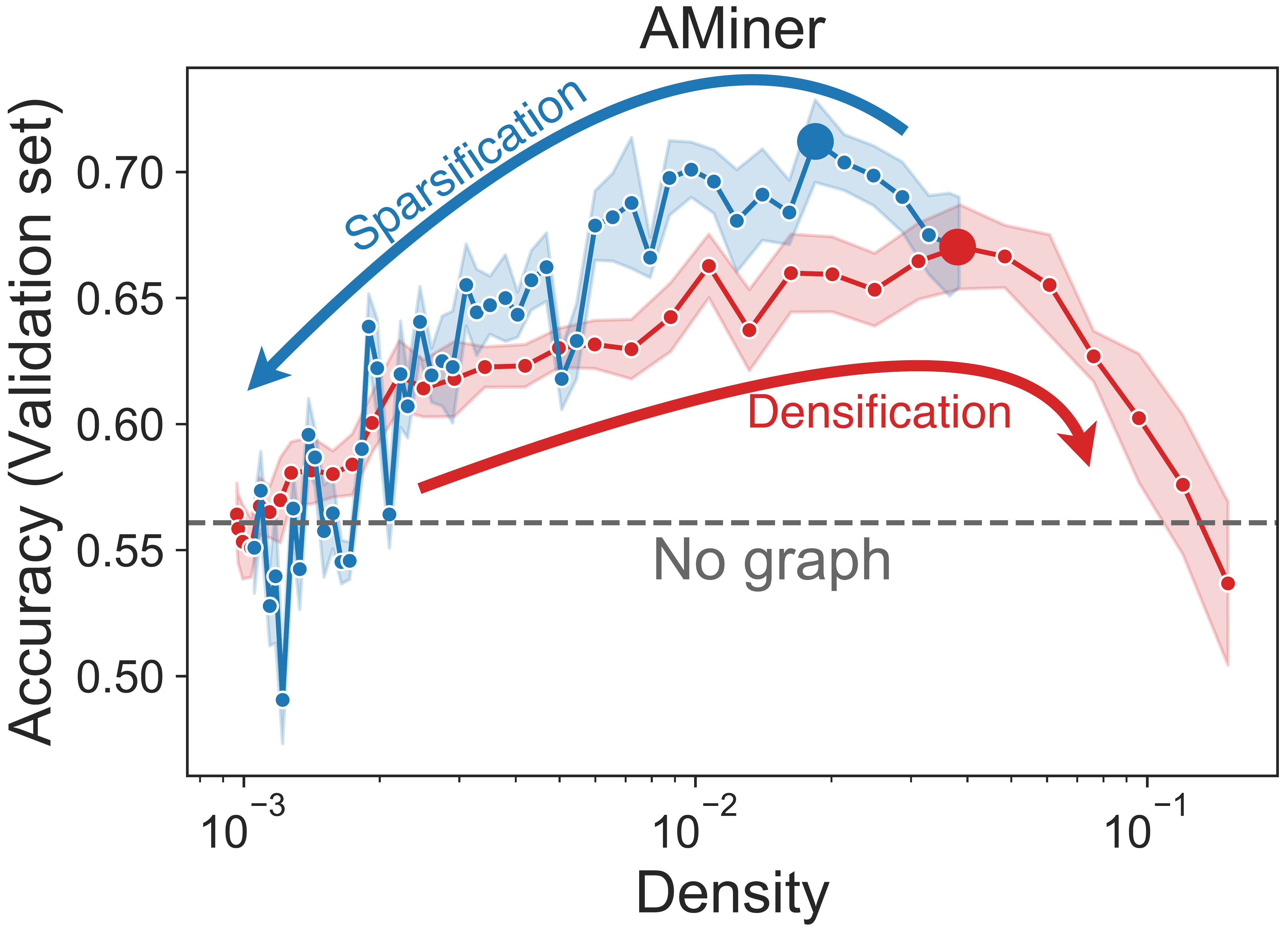}
    \label{fig:aminer_sparsification}
    \end{subfigure} \\ \bigskip
    \begin{subfigure}[c]{0.7\textwidth}
    \centering
    \caption{}
    \resizebox{\textwidth}{!}{
    \begin{tabular}{ccccc}
        \toprule
        & \multicolumn{2}{c}{\textbf{Optimized CkNN}} &  \multicolumn{2}{c}{\textbf{Sparsification of optimized CkNN}}\\
        \textbf{Data set} & $\langle \text{Degree} \rangle$ & Accuracy (Test) & $\langle \text{Degree} \rangle$ & Accuracy (Test)\\
        \midrule
        Constructive & $9.2$ & $51.1$ & $6.3$ & $51.6$\\
        Cora  & $36.7$ & $66.6$ & $36.7$ & $66.6$\\
        AMiner & $79.8$ & $61.6$ & $38.1$ & $62.5$\\
        Digits & $28.1$ & $93.4$ & $28.1$ & $93.4$\\
        FMA  & $8.0$ & $36.0$ & $8.0$ & $36.0$\\
        Cell &  $15.0$ & $84.0$ & $4.8$ & $85.0$ \\
        Segmentation & $10.3$  & $83.9$ & $8.2$ & $84.0$\\
        \midrule
        Average improvement & & (+8.3) & & (+8.7)\\
        \bottomrule
    \end{tabular}
    }
    \label{table:sparsification}
    \end{subfigure}
    \caption{\textbf{Spectral sparsification of optimized geometric graphs can further improve classification.} \\ (a) In red, the same data as in Figure~\ref{fig:aminer_CkNN_densification}, i.e., classification accuracy of GCN with CkNN graphs on AMiner data set for increasing edge density; $10$ runs with random weight initializations, shaded area is standard deviation. The large red dot indicates the optimized graph found as edges are added (densification). Starting from this optimized graph, we reduce the number of edges using the SSSA (sparsification) and record the classification accuracy on the validation set, in blue; $10$ runs with random weight initializations, shaded region is standard deviation. The large blue dot indicates the optimized sparsified graph. The gray dashed line corresponds to the classification accuracy of the MLP (no graph) on the validation set. Results for all data sets are provided in Figure~S4. 
    (b) Comparison of optimized graphs obtained through the densification and sparsification processes. The average degree of the graph ($\langle \text{Degree} \rangle$) and classification accuracy in percent on the test set are reported; averaged over $10$ runs with random weight initializations. Overall, sparsified graphs exhibit improved accuracy on the test set with lower edge density.
    }
    \label{fig:sparsification}
\end{figure}

We apply the SSSA to the optimized CkNN and select the sparsification that maximizes the classification accuracy on the validation set. Figure~\ref{fig:aminer_sparsification} shows that for the AMiner data set it is possible to improve the classification accuracy using sparser graphs obtained with SSSA. This procedure was repeated for all seven data sets (see Figure~S4). For several of our data sets, the sparsified graphs perform better on the test data with reduced edge density (see Table~\ref{table:sparsification}). 
The results of the sparsification are robust: starting the sparsification from three different highly-optimized CkNN graphs leads to similar results (see Figure~S4 and Table~S4).
Furthermore, the sparsification induces increased alignment and RCS, which correlates with the improved classification accuracy on the validation set (see Figures~S5-S6).

\section*{Discussion}
Our empirical study used data sets from different domains to show that sparse geometric graphs constructed from data features can aid classification tasks when used within the framework of GNNs.
It is worth noting that although here we have used the widely popular GCN framework to perform the classification task, other advanced GNN architectures (e.g., Deep Graph Infomax~\cite{velickovic2019deep}) could be incorporated in our pipeline as an alternative to GCN for this purpose.
In our numerics, GCN with CkNN geometric graphs display the largest improvement in classification accuracy (Table~\ref{table:results_classification}). 
This result is in line with recent work on geometric graph construction for data clustering~\cite{liu2020graph}, which showed improved behavior of CkNN over other neighborhood methods, such as kNN. CkNN graphs have been recently proposed as a consistent discrete approximation of the Laplace-Beltrami operator governing the diffusion on an underlying manifold~\cite{berry2019consistent}. Since GCN uses the graph to guide the diffusion of features to neighboring nodes, this offers a natural explanation for the good performance of CkNN under the GCN framework. Within our graph construction methods, RMST graphs use a criterion that balances neighborhood distances with non-local distances in the data set. 
While RMST outperforms graph-less methods, it does not outperform neighborhood-based methods in the examples considered here. However, RMST graphs could be appropriate for data sets where similarities based on longer paths are important.

Intuitively, geometric graphs capture the closeness (i.e., similarity) between samples in feature space, and can thus be helpful to learn and channel class labels from known samples to unseen similar samples. 
To gain further insight into why geometric graphs can improve GCN classification performance, we showed that the graph induces an increased alignment of features and ground truth, as measured by the simple measure (Equation~\eqref{eq:alignment}). The alignment correlates well with classification performance, specifically capturing the deleterious effect that over-dense graphs have on classification performance (Figure~\ref{fig:aminer_alignment}). When the graphs are over-dense, they lead to a `mean field' averaging over the whole data set which breaks the alignment---an analogous problem to the over-smoothing observed when there are too many layers in GCNs~\cite{li2018deeper,chen2020measuring}. 
We also showed that graphs with appropriate density induce increased class separability, as measured by the ratio of class separation (RCS, Equation~\eqref{eq:rcs}) derived from the GCN output activations, whereas over-sparse and over-dense graphs lead to lower class separability (Figure~\ref{fig:aminer_rcs}). 
Deviating from strictly geometric graphs, we demonstrated that spectral sparsification (SSSA) applied to the optimized CkNN graphs can be used to reduce the number of edges whilst still improving the classification performance (Figure~\ref{fig:sparsification}). Our choice of a spectral criterion for sparsification stems from the fact that the preserved Laplacian quadratic form (Equation~\eqref{eq:sparsification}) is strongly related to graph partitioning and community detection~\cite{lambiotte2014random, delvenne2010stability}. The resulting efficient graphs are thus the product of a mixed process: a geometric graph provides a local similarity neighborhood which is further sharpened using global graph properties captured by the Laplacian spectrum.

Methods that leverage graphs in data analysis have a long history~\cite{lauritzen1996graphical}, 
and have been recently considered in conjunction with deep learning algorithms.
Ref.~\cite{franceschi2019learning} proposed a novel method that jointly learns graph structure and the parameters of a GNN by solving a bilevel program to obtain a discrete probability distribution on the edges of the graph.
We have compared our method with the one proposed in Ref.~\cite{franceschi2019learning}. Our results are summarized in Table~S5 and indicate that our proposed method achieves, on average, classification accuracy comparable to Ref.~\cite{franceschi2019learning}, yet with a significantly smaller number of parameters, thus simplifying the training and reducing the inclination to overfitting. Table~S5 also includes the results~\cite{qian2021quantifying} obtained by applying GCN to data sets that contain a graph as an additional source of information (i.e., the citation networks for Cora and AMiner). The improved accuracy of GCN with these original graphs stems from the additional information the graphs contain, beyond what is present in the features alone. Specifically, the original graphs for Cora and AMiner collate citations between scientific articles, which encode additional information about the similarity between articles not captured by the features (i.e., the text embedding vectors) of the articles themselves.
Another recent method constructed a local neighborhood graph as part of convolution-based classifiers~\cite{wang2019dynamic}. 
Our work, in contrast, focuses on graph-theoretical measures~\cite{liu2020graph}, by exploring different graph constructions and the importance of edge density and spectral content for classification, and characterizing the effect of graphs through geometric notions of separability and subspace alignment~\cite{qian2021quantifying}.

In our numerical experiments, feature-derived geometric graphs appear to be most useful when the data is high-dimensional, noisy, and co-linearity is present in the features. In particular, GCN with optimized graphs outperforms the graph-less methods in all our data sets except `Segmentation'. 
All our data sets are high-dimensional without feature engineering except `Segmentation', a data set with $19$ engineered features specifically optimized for classification---this is the set-up where SVM and RF are expected to work well. However, even in that case, we note that the featured-derived graphs still improve the classification performance with respect to MLP, indicating that the graphs help filter out feature similarities that can obscure the action of MLP.

Beyond the potential to improve performance, using graphs to aid classification changes the paradigm from supervised to semi-supervised learning.
Supervised methods, e.g., MLP, perform inductive learning, whereas graph-based semi-supervised learning can be either transductive or inductive. GCNs belong to transductive learning, since the graph of the \emph{full} data set is used for the training. Therefore, whilst potentially advantageous, the use of GCNs can also restrict the generalizability to new samples. In many applications such a requirement does not impose severe restrictions, but graph-based methods can still be adapted to classify new data without the need to recompute the model. For instance, one could predict the class label of a new sample directly from the output activations $Z$ of the closest samples in the original set, or using more elaborate diffusion-based schemes~\cite{peach2020semi}.

Our proposed pipeline also shares common ground with some of the most successful clustering methods developed for single-cell genomics data sets. For example, Seurat~\cite{satija2015spatial} uses Louvain modularity optimization to perform community detection on a kNN graph constructed from the top principal components of data. Similarly, other methods for graph-based clustering have been introduced using multiscale extensions of the Louvain algorithm in the framework of Markov Stability~\cite{liu2020graph}. Although classification and clustering are different learning tasks, we have carried out a comparison between our proposed method (CkNN+GCN) and two Louvain-based clustering methods (Seurat and a straightforward kNN+Louvain clustering). After optimizing each method using the training and validation sets, we computed the assignment it produces on the test set, and compared it to the ground truth classes (see Supplemental Information 2). The quality of the assignments (evaluated with the Adjusted Rand Index and Normalized Mutual Information) presented in Table~S6 indicates that, on average across our data sets, our proposed method performs better than Seurat's approach.

Our study opens several avenues for future work. Here we explored graph construction based on geometry; it will be interesting to consider graph construction paradigms that incorporate other criteria, e.g., small-worlds~\cite{watts1998collective}, graph expanders~\cite{hoory2006expander}, or entangled networks~\cite{donetti2005entangled}, among others. Similarly, although we showed that spectral sparsification~\cite{spielman2011spectral} is a good choice to improve efficiency, other graph sparsification paradigms, e.g., cut sparsification~\cite{fung2019general}, might also be useful to achieve efficient graphs for classification. Here we have adopted the Euclidean distance as a simple metric to base our geometric graph construction. However, other metrics could be used in our pipeline and could be indeed more appropriate for different types of data. Investigating the effect of different distance metrics (such as the Manhattan distance, cosine similarity, or distances in transformed spaces such as PCA or other projections) would be an important question for future research.

\section*{Experimental Procedures}
\subsection*{Resource Availability}
\subsubsection*{Lead Contact}
Further information and requests for resources should be directed to and will be fulfilled by the Lead Contact, Mauricio Barahona (m.barahona@imperial.ac.uk).
\subsubsection*{Materials Availability}
This study did not generate any unique reagents.
\subsubsection*{Data and Code Availability}
The data sets for graph construction can be found at~\url{https://github.com/haczqyf/ggc}. The code for graph construction can be found at~\url{https://github.com/haczqyf/ggc}. The sources for other code (e.g., GCN and graph sparsification) are described in the Supplemental Information 3. The algorithm complexity of our proposed pipeline has been discussed in the Supplemental Information 4. The run time and memory requirements have been described in the Supplemental Information 5.

\subsection*{Methods}
\subsubsection*{Graph Construction\label{sec:graph_construction}}
Let $X_{i}$ be the $F$-dimensional feature vector (L1-normalized) of the $i$-th sample of our data set with $N$ samples. The pairwise dissimilarity between samples $i$ and $j$ is taken to be the Euclidean distance:
\begin{equation}
    d(i,j) = \left\|X_{i} - X_{j}\right\|_{2}.
\end{equation}
The distance matrix of all samples $D \in R^{N \times N}$ where $D_{ij}=d(i,j)$ is then used to construct unweighted and undirected graphs based on different heuristics.  
To guarantee connectedness over the data set, we first construct the MST. The MST is obtained from the Euclidean distance matrix $D$ using the Kruskal algorithm, and contains the $N-1$ edges that connect all the nodes (samples) in the graph with minimal sum of edge weights (distances). Once the weighted MST is obtained, we ignore the edge weights, as is also done for all other graphs in the paper. Thus the resulting graphs are undirected and unweighted.
The 0-1 adjacency matrix of the MST is denoted by $A^{\text{MST}}$. We then add edges to the MST based on two types of criteria: (i) local neighborhoods, or (ii) balancing local and global distances.

\paragraph*{Methods Based on Local Neighborhoods: Nearest Neighbors} 

The objective of neighborhood-based methods is to construct a sparse graph by connecting two samples if they are local neighbors, as determined by $d(i,j)$. 

The simplest such algorithm is kNN. A kNN graph has an edge between two samples $i$ and $j$ if \textit{one} of them belongs to the $k$-nearest neighbors of the other. The adjacency matrix $A^{\text{kNN}}\in R^{N \times N}$ of a kNN graph is defined by:
\begin{equation}
  A^{\text{kNN}}_{i,j} = 
  \begin{cases} 
  1       & \text{if } d(i,j) \leq d(i, i_{k}) \text{ or }  d(i,j) \leq d(j, j_{k}) \\
  0       & \text{otherwise}
  \end{cases}
\end{equation}
where $i_{k}$ and $j_{k}$ represent the $k$-th nearest neighbors of samples $i$ and $j$, respectively.

Although widely used, kNN has limitations. Perhaps most importantly, kNN graphs can have highly heterogeneous degree distributions and often contain hubs, i.e., samples with high number of connections, since kNN greedily connects two samples as long as one of them belongs to the other's $k$-nearest neighbors. It has been suggested that the presence of hubs in kNN graphs is particularly severe when the samples are high-dimensional~\cite{radovanovic2010hubs}. It has been observed that hubs tend to deteriorate the classification accuracy of semi-supervised learning~\cite{ozaki2011using}. 

To overcome this limitation, the MkNN algorithm, a variant of kNN, was proposed~\cite{ozaki2011using}. In an MkNN graph an edge is established between samples $i$ and $j$ if \textit{each} of them belongs to the other's $k$-nearest neighbors. The adjacency matrix $A^{\text{MkNN}}\in R^{N \times N}$ of the MkNN graph is defined by:
\begin{equation}
  A^{\text{MkNN}}_{i,j} = 
  \begin{cases} 
  1       & \text{if } d(i,j) \leq d(i, i_{k}) \text{ and }  d(i,j) \leq d(j, j_{k}) \\
  0       & \text{otherwise }
  \end{cases}
\end{equation}
Note that the MkNN algorithm guarantees that the degrees of all samples are bounded by $k$. Therefore, MkNN reduces the presence of hubs when $k$ is adequately small.

Another limitation of kNN is its lack of flexibility to provide a useful, stable graph when the data is not uniformly sampled over the underlying space, which is often the case in practice~\cite{liu2020graph}. In such situations, it is difficult to find a single value of $k$ that can accommodate the disparate levels of sampling density across the data, since the kNN graph will connect samples with very disparate levels of similarity depending on the region of the sample space (i.e., in densely sampled regions, the graph only connects data points that are very similar, whereas in poorly sampled regions, the graph connects data samples that can be quite dissimilar). This disparity biases the training of the GCN. The non-uniformity of the data distribution thus makes it difficult to tune a unique $k$ parameter that is appropriate across the whole data set. If the value of $k$ is too small, the graph is dominated by local noise, and fails to provide consistent information to improve the GCN training. If the value of $k$ is large, the resulting graph is over-connected and leads GCN to degraded accuracy, close to mean-field classification. Hence, when the sampling is not homogeneous, standard kNN graphs can be sub-optimal in capturing the underlying data structure with a view to improved learning.

CkNN~\cite{berry2019consistent} has recently been introduced to address this limitation by allowing an adjusted local density. The adjacency matrix $A^{\text{CkNN}}\in R^{N \times N}$ associated with a CkNN graph is defined by:
\begin{equation}
  A^{\text{CkNN}}_{i,j} = 
  \begin{cases} 
  1       & \text{if } d(i,j) < \delta\sqrt{d(i, i_{k})d(j, j_{k})} \\
  0       & \text{otherwise }
  \end{cases}
\end{equation}
where the parameter $\delta >0$ regulates the density of the graph. For a fixed $k$, the larger $\delta$ is, the denser the CkNN graph becomes. Ref.~\cite{berry2019consistent} shows that the CkNN graph captures the geometric features of the data set with the additional consistency that the unnormalized Laplacian of the CkNN graph converges spectrally to the Laplace-Beltrami operator in the limit of large data. In this work, we fix $\delta=1$ and vary $k$ so that CkNN can be compared with kNN and MkNN, as suggested in Ref.~\cite{liu2020graph}.

All these three methods capture the geometry of local neighborhoods, with global connectivity guaranteed by the MST.

\paragraph*{Balancing Local and Global Distances: Relaxed Minimum Spanning Tree}

Alternatively, other graph constructions attempt to balance the local geometry with a measure of global geometry extracted from the full data set. In recent years, several algorithms have been introduced to explore global properties of the data using the MST~\cite{beguerisse2013finding,liu2020graph}. Here, we focus on the RMST~\cite{beguerisse2013finding}, which considers the largest distance $d^{\text{max}}_{\text{MST-path}(i,j)}$ encountered along the unique MST-path between $i$ and $j$. If $d^{\text{max}}_{\text{MST-path}(i,j)}$ is substantially smaller than $d(i,j)$, RMST discards the direct link between $i$ and $j$, recognizing the multi-step MST-path as a good model to capture the similarity between them. If, on the other hand, $d(i,j)$ is comparable to $d^{\text{max}}_{\text{MST-path}(i,j)}$, the MST-path does not provide a good model, and RMST adds the direct link between $i$ and $j$. The adjacency matrix $A^{\text{RMST}}\in R^{N \times N}$ associated with a RMST graph is defined by:
\begin{equation}
  A^{\text{RMST}}_{i,j} = 
  \begin{cases} 
  1       & \text{if } d(i,j) < d^{\text{max}}_{\text{MST-path}(i,j)} + \gamma(d(i,i_{k})+d(j,j_{k})) \\
  0       & \text{otherwise }
  \end{cases}
\end{equation}
where $\gamma \geq 0$ is the density parameter, and $d(i,i_{k})$ and $d(j,j_{k})$ approximate the local distribution of samples around $i$ and $j$, respectively, as the distance to their $k$-th nearest neighbor~\cite{zemel2005proximity}. Here, we fix $k=1$ and vary $\gamma$ to change the edge density, as in Ref.~\cite{liu2020graph}. 

\subsubsection*{Spectral Graph Sparsification}
The graph construction methods above can be thought of as a \textit{graph densification}, in which the starting point is the MST over the $N$ samples and an edge is added between two samples $i$ and $j$ if the distance $d(i,j)$ meets a defined criterion. 
Graph sparsification operates in the opposite direction: starting from a given graph, the goal is to obtain a sparser graph that approximates the original graph so that it can be used, e.g., in numerical computations, without introducing too much error. Sparsified graphs are more efficient for both numerical computation and data storage~\cite{spielman2011spectral}. Here, we focus on spectral graph sparsification~\cite{spielman2011spectral}, and apply the seminal SSSA proposed in Ref.~\cite{spielman2011graph}. SSSA obtains a spectral approximation of the given graph that satisfies the following criterion:
\begin{equation}
    (1-\sigma) \,  x^{\mathrm{T}} L x \leq x^{\mathrm{T}} \widetilde{L} x  \leq (1+\sigma) \,  x^{\mathrm{T}} L x
    \label{eq:sparsification}
\end{equation}
where $x\in R^{N \times 1}$ is a node vector, and $L$ and $\widetilde{L}$ are the Laplacian matrices of the original and sparsified graphs, respectively. For each data set, we obtain increasingly sparse versions of the optimized geometric graph computed above by scanning over $50$ values of the sparsity parameter $\sigma$ between $1/N$ and $1$. At each value of $\sigma$, we run the GCN classifier $10$ times starting from random weight initializations and compute the average accuracy over the validation set. We then select the graph with highest accuracy and maximum sparsity. If sparsification does not improve performance on the test set, we report the unsparsified graph as optimal (e.g., in Cora, Digits and FMA in Figure~\ref{table:sparsification}).

\subsubsection*{Graph Convolutional Networks}
GNNs, a new class of deep learning algorithms, have been recently proposed to analyze graph-structured data. Here, we focus on GCNs~\cite{kipf2017semi} and their application to semi-supervised learning. Each sample $i$ is characterized by an $F$-dimensional feature vector $X_{i} \in R^{1 \times F}$, which is arranged as a row of the feature matrix $X \in R^{N \times F}$. In addition, the $N$ samples are associated with a graph $\mathcal{G}$ where the samples are the nodes and edges represent relational (symmetric) information. The graph is described by an adjacency matrix $A \in R^{N \times N}$. Each sample is also associated with one of $C$ classes, which is encoded into a $0$-$1$ membership matrix $Y \in R^{N \times C}$. GCNs train a model using the full feature matrix $X$, the adjacency matrix $A$ of the full graph, and a small subset of ground truth labels, i.e., a few rows of $Y$. The obtained model is then used to predict the class of unlabeled nodes and evaluate the classification performance by comparing inferred labels with their ground truth labels.

Our study applies the two-layer GCN proposed in Ref.~\cite{kipf2017semi}. Given the feature matrix $X$ and the (undirected) adjacency matrix $A$ of the graph $\mathcal{G}$,
the propagation rule is given by:
\begin{equation}
Z = f(X,A)
 = \mathrm{softmax}(\widehat{A}~\mathrm{ReLU}(\widehat{A}XW^{0})~W^{1}),\\
\label{eq:two_layers_rule}
\end{equation}
where $W^{0}$ and $W^{1}$ are the weights connecting the layers of the GCN. The graph is encoded in
$\widehat{A}=\widetilde{D}^{-1/2}(A + I_{N})\widetilde{D}^{-1/2}$, where $I_{N}$ is the identity matrix, and $\widetilde{D}$ is a diagonal matrix with $\widetilde{D}_{ii} = 1 + \sum_{j}A_{ij}$. The $\mathrm{softmax}$ and $\mathrm{ReLU}$ are threshold activation functions:
\begin{align}
\mathrm{ReLU}(x)_{i}=\max(x_{i},0)  \\
\mathrm{softmax}(x)_{i} = \frac{\exp(x_{i})}{\sum_{j} \exp(x_{j})}
\end{align}
where $x$ is a vector. The cross-entropy error over all labeled samples is:
\begin{equation}
        \mathcal{L} = -\displaystyle\sum_{l\in \mathbb{Y}_{L}}\displaystyle\sum_{c=1}^{C} Y_{lc} \, \ln{Z_{lc}},
    \label{eq:cross_entropy_error}
\end{equation}
where $\mathbb{Y}_{L}$ is the set of nodes that have labels (i.e., the training set). The  weights of the neural network ($W^{0}$ and $W^{1}$) are trained using gradient descent to minimize the loss~$\mathcal{L}$. 

In our case, the classification is based solely on information obtained from the features since the graphs are also feature-derived. 

\paragraph*{GCN Architecture, Hyperparameters and Implementation}
We use the GCN implementation provided by the authors of Ref.~\cite{kipf2017semi}, and follow closely the experimental setup in Refs.~\cite{kipf2017semi,qian2021quantifying}. We use a two-layer GCN with $2,000$ epochs (training iterations); learning rate of $0.01$; and early stopping with a window size of $200$. Other hyperparameters are: dropout rate: $0.5$; L2 regularization: $5\times 10^{-4}$; number of hidden units: $16$. The weights are initialized as described in Ref.~\cite{glorot2010understanding}, and the input feature vectors are L1 row-normalized. We choose the same data set splits as in Ref.~\cite{qian2021quantifying} with $5\%$ of samples as training set, $10\%$ of samples as validation set and the remaining $85\%$ as test set (see Table~S1). The samples in the training set are evenly distributed across classes.

\subsubsection*{Graph-less Classification Methods\label{sec:baselines}}
For comparison, we consider four graph-less classification methods: 
(i) MLP, which is equivalent to GCN with no graph~\cite{kipf2017semi,qian2021quantifying}; 
(ii) kNNC based on the plurality vote of its $k$-nearest neighbors; 
(iii) SVM with Radial Basis Function kernel; 
and (iv) RF. 
We use the Scikit-learn~\cite{pedregosa2011scikit} implementation for kNNC, SVM and RF.
For each method and each data set, we use the validation set to optimize the following hyperparameters: number of neighbors (kNNC); regularization parameter (SVM); maximum depth (RF). All other hyperparameters are set as default in Scikit-learn.
We compare the graph-less methods against the MLP = GCN (No graph), which is used as the reference baseline.

\section*{Acknowledgements}
Y.Q. acknowledges financial support from the China Scholarship Council Program (No. 201706020176). P.E. is funded by the National Institute for Health Research (NIHR) Imperial Biomedical Research Centre (BRC) under Grant NIHR-BRC-P68711 and acknowledges partial support from Engineering and Physical Sciences Research Council (EPSRC) under Grant EP/N014529/1 supporting the EPSRC Centre for Mathematics of Precision Healthcare. M.B. acknowledges support from EPSRC under Grant EP/N014529/1 supporting the EPSRC Centre for Mathematics of Precision Healthcare.

\section*{Author Contributions}
Y.Q., P.E., P.P., and M.B. designed research; Y.Q., P.E., P.P., and M.B. performed research; Y.Q. analyzed data; and Y.Q., P.E., P.P., and M.B. wrote the paper.

\section*{Declaration of Interests}
The authors declare no competing interest.


\end{document}


\title{\textbf{SUPPLEMENTAL INFORMATION:\\``Geometric Graphs from Data to Aid Classification Tasks with Graph Convolutional Networks''}}

\author[1]{Yifan Qian}
\author[2,3]{Paul Expert}
\author[1,*]{Pietro Panzarasa}
\author[4,5,*]{Mauricio Barahona}
\affil[1]{School of Business and Management, Queen Mary University of London, London, UK}
\affil[2]{Global Digital Health Unit, School of Public Health, Imperial College London, London, UK}
\affil[3]{World Research Hub Initiative, Tokyo Institute of Technology, Tokyo, Japan}
\affil[4]{Department of Mathematics, Imperial College London, London, UK}
\affil[5]{Lead Contact}
\affil[*]{Correspondence: m.barahona@imperial.ac.uk (M.B.), p.panzarasa@qmul.ac.uk (P.P.)}
\date{}

\maketitle

\newpage

\section*{Supplemental Information 1: Data sets}
We use seven data sets collected from various sources. We provide the data sets at \url{https://github.com/haczqyf/ggc/tree/master/ggc/data}. The data set statistics are summarized in Supplemental Table~\ref{table:dataset_statistics}.

\begin{enumerate}
\item \textit{Constructive}~\cite{qian2021quantifying} is a synthetic data set generated by a stochastic block model that reproduces the ground truth structure with some noise. Each ground truth cluster is associated with $50$ features with a probability of $p_{\rm{in}}=0.07$ equal to $1$. Each sample also has a probability of $p_{\rm{out}}=0.007$ of possessing each feature characterizing other clusters.

\item We consider two data sets with text documents: \textit{Cora}~\cite{sen2008collective,qian2021quantifying} and \textit{AMiner}~\cite{qian2017citation,tang2008arnetminer}. In Cora and AMiner, the samples are scientific papers where each paper is associated with a high-dimensional bag-of-words feature vector extracted from the paper content. Each sample has a class label indicating its scientific field.

\item \textit{Digits} is a handwritten digits data set. Each sample is a 8x8 image of a digit. This is one of the benchmark data sets for classification in Scikit-learn~\cite{pedregosa2011scikit}.

\item \textit{FMA}: The original data set~\cite{franceschi2019learning,defferrard2017fma} contains $140$ audio features extracted from $7,994$ music tracks. We use this data set to address the problem of genre classification. The original data set in ref.~\cite{franceschi2019learning} contains $8$ genres. We sample randomly $2,000$ music tracks ($250$ for each genre) to produce our data set.

\item \textit{Cell}: This is a data set of brain cell types from autism. The original data set~\cite{velmeshev2019single} contains the gene expression values ($\log_2$ transformed 10x UMI counts from cellranger) of $104,599$ single cells from brains of control individuals and of patients with autism, where each cell (sample) is characterized by the expression level of $36,501$ genes (features). The full data set contains cells from $17$ cell types (categories). To produce our data set, we sample randomly $2,000$ cells from $10$ cell types ($200$ cells for each type) and select as our features the expression level of the top $500$ most highly variable genes across the $2,000$ cells in our sample.

\item \textit{Segmentation}: This is an image segmentation data set, which is provided at UCI machine learning repository~\cite{Dua:2019} at~\url{https://archive.ics.uci.edu/ml/datasets/Image+Segmentation}. Each sample represents an image described by $19$ high-level and man-crafted numeric-valued attributes.

\end{enumerate}

\section*{Supplemental Information 2: Comparison with Seurat clustering}
Single-cell clustering is indeed an area where some of the gold standard methods are based on applying community detection to graphs derived from cell features (e.g., gene expression levels) by using the Louvain algorithm to maximize modularity. 
Seurat~\cite{satija2015spatial} is such a graph-based clustering approach, where a kNN graph is constructed from the PCA decomposition of the original feature vectors, and the obtained graph is then partitioned into communities (corresponding to cell types) by Louvain modularity maximization. \\~\\ 
It is important to remark that there is a fundamental distinction between the Seurat setting and our work. While our method (CkNN+GCN) addresses a classification problem (supervised setting, in which some class labels are known as ground truths and used in the training), Seurat solves a clustering problem (unsupervised setting, in which there are no class labels available on which to train). Clustering aims to group similar samples together and dissimilar samples into distinct groups based on the similarity between their features~\cite{liu2020graph}.
Seurat clustering involves three steps: (i) compute the principal components of the feature vectors, and select the top $T$ principal components based on a choice of $p$, the ratio of explained variance to total variance; (ii) construct a kNN graph based on the Euclidean distance between the vectors defined by the top $T$ principal components of each sample; and (iii) perform community detection on the kNN graph using Louvain modularity maximization. In this process, several hyperparameters are chosen, including the ratio $p$, which determines the number of principal components in step (i), and the number of neighbors $k$ in the kNN  graph in step (ii).
The final result of Seurat is a partition of the data set into clusters (`graph communities') derived intrinsically from properties of the data. \\~\\
Our method (CkNN+GCN), on the other hand, attempts a classification task where we leverage both the features and a feature-derived CkNN graph of appropriate edge density to train the weights of a GCN in order to maximize its classification power. Our use of GCN and CkNN is distinctive in this setting, as is the optimization of the edge density of the graph to maximize the quality of the classification.  
Given that the objectives of Seurat and our method are different, it is not straightforward to compare both approaches, but we have produced a setting to compare both methods. In particular, we have devised a comparison between our CkNN+GCN method and two Louvain-based clustering methods: Seurat (PCA+kNN+Louvain) and a simpler application of Louvain to a kNN graph of features (kNN+Louvain) without applying PCA in the first step. These three methods are compared to a simple kNN classifier (kNNC). \\~\\
To compare the methods, we use the labels in the training and validation sets (defined as above in our CkNN+GCN experiments) as ground truths, and tune the hyperparameters $k$ and $p$ ($k$ is grid-searched over $[2,4,8,16,32,64]$ and $p$ is grid-searched over $[0.5,0.6,0.7,0.8,0.9]$) to maximize the similarity between the obtained clusters and the ground truth partitions. Once the hyperparameters have been optimized, we then use each method to cluster the data and we compute the quality of the clustering against the test set. To evaluate the quality of the clustering we use two standard measures: the Adjusted Rand Index (ARI) and the Normalized Mutual Information (NMI). Both of these measures are normalized between $0$ (random assignment) and $1$ (perfect agreement), with higher values signifying better assignment. Our results are presented in Table~\ref{table:compare_clustering_ARI_NMI}.
Our results show that our method (CkNN+GCN) performs better on average than both Louvain-based clustering methods on our data sets. Yet CkNN+GCN does not always outperform the other methods; in particular, Seurat is the best on the Cell data set. 
This might reflect particularities of the Cell data set, which contains high-dimensional vectors with high levels of noise that might benefit from the effective dimensionality reduction and filtering that PCA enforces. On the other hand, CkNN+GCN has been kept as a broad-purpose method, i.e., not optimized for a particular type of data. For instance, we use default values for some GCN hyperparameters (learning rate, number of hidden units, drop out rate) without optimizing them on each data set. The aim is to provide robust outcomes across diverse data sets, as shown in Table~\ref{table:compare_clustering_ARI_NMI}. Hence, there is room to potentially optimize our method (CkNN+GCN) specifically for single-cell genomics, but we feel this falls beyond the scope of our current work, and will be investigated in future research. \\~\\  
Still, we would like to remark that clustering and classification algorithms are not directly comparable since they have different objectives and learning contexts. Nonetheless, we hope that our additional experiments provide some insight into the comparison.

\section*{Supplemental Information 3: Code availability}
We provide the data sets and code for geometric graph construction at~\url{https://github.com/haczqyf/ggc}. The code for Graph Convolutional Networks (GCNs) is provided by the authors of~\cite{kipf2017semi} at \url{https://github.com/tkipf/gcn}. The code for kNN Classification (kNNC), Support Vector Machine (SVM) and Random Forest (RF) can be found at~\url{https://scikit-learn.org/stable/} from scikit-learn~\cite{pedregosa2011scikit}. The code for Spielman-Srivastava sparsification algorithm (SSSA) is available at \url{https://epfl-lts2.github.io/gspbox-html/doc/utils/gsp_graph_sparsify.html} from Graph Signal Processing Toolbox~\cite{perraudin2014gspbox}.

\section*{Supplemental Information 4: Algorithm complexity}
For a graph $\mathcal{G}=(\mathcal{V}, \mathcal{E})$ with $N$ nodes $v_i \in \mathcal{V}$ and $|\mathcal{E}|$ edges $(v_i, v_j) \in  \mathcal{E}$, the time complexity for GCN, i.e., to evaluate Equation~(9), is $O(|\mathcal{E}| F H C)$~\cite{kipf2017semi}, where $|\mathcal{E}|$ is the number of graph edges, $F$ is the dimension of the feature space, $H$ is the number of units in the hidden layer and $C$ is the number of classes in the ground truth. Hence the computational complexity for GCN is linear in the number of graph edges. 
For the geometric graph construction, a brute force approach to compute exactly a geometric graph (i.e., the kNN-type graphs) has time complexity $O(FN^{2})$. However, fast approximate kNN graph algorithms were proposed to reduce this time complexity. We mention two examples: (i) Ref.~\cite{chen2009fast} proposes an algorithm with complexity $O(FN^{t})$ with $1<t<2$, and (ii) Ref.~\cite{andoni2006near} proposes an algorithm that uses locality sensitive hashing, which has complexity $O\left(FN^{1 / c^{2}+o(1)}\right)$ where $c=1+\epsilon>1$. For a thorough list of approximate kNN algorithms, see https://github.com/stephenleo/adventures-with-ann/.
Regarding spectral sparsification, the algorithm is nearly linear with time complexity $\tilde{O}(|\mathcal{E}|)$~\cite{spielman2011graph}, 
where the $\tilde{O}$ notation ignores logarithmic factors. Finally, for the MST construction, we use the Kruskal algorithm implemented in~Scipy with time complexity $O(|\mathcal{E}|{\rm{log}}N)$.

\section*{Supplemental Information 5: Run time and memory requirements}
To give a sense of run times and memory requirements for our algorithm, we summarize briefly the numbers for the Cora data set, which presents the worst-case run times and storage requirements among our seven examples. Indeed, we find that Cora has the longest run times, consistent with the algorithmic complexity in Supplemental Information 4, since Cora has the largest number of nodes and highest dimensions. For graph construction, creating and storing in disk all the graphs during the optimization of the hyperparameter takes around $13$ hours with a maximum used memory around 3G. However, our algorithm can be further optimized since the graphs do not have to be stored and could be created and used on the fly to save memory usage and access time. Furthermore, over-dense graphs could be avoided altogether since the optimized graphs usually are relatively sparse. Indeed, we find that the graphs with optimal accuracy have densities on the order of $0.005 - 0.05$ of the total number of possible edges (see Table~S3 in ``Supplemental Tables and Figures''), and for densities above $\sim 0.1$ the accuracy drops below the accuracy of an MLP. For higher densities, the accuracy consistently degrades towards the random assignment limit. Therefore the grid search of the hyperparameter can be restricted to low density graphs, and dense graphs do not have to be stored or computed. The search for the optimal hyperparameter can be further aided with a bisection scheme and could be parallelized to improve the efficiency of the optimization. \\~\\
For a thorough description of the complexity of the different blocks of our algorithm (GCN, kNN and MST graph constructions, and spectral sparsification) see Supplemental Information 4.     For each value of the hyperparameter, we run a GCN $10$ times from $10$ random initializations. The cost of each GCN is moderate: the complexity of GCN scales nearly linearly with the number of edges of the graph. The cost of constructing kNN-type graphs (originally of $O(N^2)$) can also be reduced to nearly linear in the number of nodes with approximation algorithms. Sparsification is also nearly linear, as shown by Spielman. Hence the methodology has the potential to be applied to relatively large graphs with further code optimization. For instance, each GCN for Cora takes typically less than $7$ minutes for relatively sparse graphs ($k\leq 200$), and each graph sparsification takes less than $2$ minutes. \\~\\
Comparing to the Louvain-based methods, there is the same complexity for the kNN graph construction, whereas the run time complexity of Louvain optimization is $O(N{\rm{log}^{2}}N)$. For Seurat, there is the additional cost of performing PCA to extract the top $T$ principal components, with complexity $O(N^{2}T)$ (inherited from randomized SVD). Thus, the run time complexity and memory requirements of the Louvain-based methods are comparable to those of our method.

\section*{Supplemental Tables and Figures}

\def\sym#1{\ifmmode^{#1}\else\(^{#1}\)\fi}

\begin{table}[H]
\centering
\caption{Summary statistics of the data sets in our study.}
\resizebox{0.75\textwidth}{!}{
\begin{tabular}{ccccccccc}
\toprule
\textbf{Data sets} & \textbf{Type} & \textbf{Samples ($N$)} &
\textbf{Features ($F$)} & \textbf{Classes ($C$)}& \textbf{Train$/$Validation$/$Test} \\
\midrule
Constructive & Stochastic block model & $1,000$
& $500$ & $10$ & $50/100/850$\\
Cora & Text (Bag-of-words) & $2,485$
& $1,433$ & $7$& $119/253/2,113$\\
AMiner & Text (Bag-of-words) & $2,072$
& $500$ & $7$& $98/212/1,762$\\
Digits & Images (Grayscale pixels) & $1,797$
& $64$ & $10$& $80/189/1,528$ \\
FMA (songs) & Music track features & $2,000$
& $140$ & $8$& $96/204/1,700$ \\
Brain cell types & Single-cell transcriptomics  & $2,000$ & $500$ & $10$& $100/200/1,700$\\
Segmentation & Image features  & $2,310$
& $19$ & $7$ & $112/234/1,964$\\
\bottomrule
\end{tabular}}
\label{table:dataset_statistics}
\end{table}

\begin{table}[h]
\centering
\caption{
Classification accuracy (in percent) on the test set (average and standard deviation over $10$ runs with random initializations) for $7$ data sets with $8$ classifiers (four graph-less methods; GCN with four graph constructions).
}
\resizebox{0.75\textwidth}{!}{
    \begin{tabular}{c|ccccccc}
        \toprule
        \textbf{Classifier} & \textbf{Constructive} & \textbf{Cora} & \textbf{AMiner} & \textbf{Digits} & \textbf{FMA} & \textbf{Cell}  & \textbf{Segmentation}  \\
        \midrule
        MLP = GCN (No graph) &  42.1 $\pm$ 1.2 &  54.2 $\pm$ 1.7  &  54.4 $\pm$ 1.1 & 82.0 $\pm$ 1.3 & 34.3 $\pm$ 0.8 & 79.5 $\pm$ 3.0 &72.0 $\pm$ 2.4 \\
        kNNC  &  31.4 $\pm$ 0.0 & 38.2 $\pm$ 0.0 & 28.0 $\pm$ 0.0 & 88.3 $\pm$ 0.0 & 30.6 $\pm$ 0.0 & 58.7 $\pm$ 0.0& 68.8 $\pm$ 0.0 \\
        SVM &  40.0 $\pm$ 0.0&55.9 $\pm$ 0.0  &51.4 $\pm$ 0.0  &87.7 $\pm$ 0.0  &35.3 $\pm$ 0.0  &81.5 $\pm$ 0.0 &87.7  $\pm$ 0.0\\
        RF & 36.3 $\pm$ 1.0 & 56.1 $\pm$ 1.2 &47.7 $\pm$ 1.5  &83.0 $\pm$ 0.5  &33.0 $\pm$ 0.9  &88.0 $\pm$ 0.7  &88.8 $\pm$ 0.7\\
        \midrule
        GCN (kNN) & 53.9 $\pm$ 0.9 & 66.4 $\pm$ 0.6 & 59.2 $\pm$ 1.3  & 92.0 $\pm$ 0.4 &35.6 $\pm$ 1.0 &83.8 $\pm$ 1.6 & 83.5 $\pm$ 0.7\\ 
        GCN (MkNN) & 45.2 $\pm$ 1.6 & 64.1 $\pm$ 0.4&61.8 $\pm$ 0.8 & 93.2 $\pm$ 0.3&35.6 $\pm$ 0.7 & 84.0 $\pm$ 2.0 & 83.0 $\pm$ 0.6\\ 
        GCN (CkNN) & 51.1 $\pm$ 1.3 & 66.6 $\pm$ 0.4 & 61.6 $\pm$ 0.8  & 93.4 $\pm$ 0.3 &  36.0 $\pm$ 0.8& 84.0 $\pm$ 2.1& 83.9 $\pm$ 0.6\\
        GCN (RMST) & 45.9 $\pm$ 1.5 & 64.8 $\pm$ 0.6 & 61.5 $\pm$ 1.3 &  89.3 $\pm$ 0.5 & 35.4 $\pm$ 0.7 & 84.9 $\pm$ 1.1 & 83.0 $\pm$ 1.6\\
        \bottomrule
\end{tabular}}
\end{table}

\begin{table}[H]
\centering
\caption{Selected density parameters and density of constructed graphs in the graph densification process (Supplemental Figure~\ref{fig:densification_density})}
\label{table:summary_parameters_optimized_graph_construction}
\resizebox{0.75\textwidth}{!}{
\begin{tabular}{ccccccccc}
\toprule
& \multicolumn{2}{c}{\textbf{kNN}} & \multicolumn{2}{c}{\textbf{MkNN}}  & \multicolumn{2}{c}{\textbf{CkNN} ($\delta=1$)} & \multicolumn{2}{c}{\textbf{RMST} ($k=1$)}\\
\textbf{Data set} & $k^{*}$ & Density & $k^{*}$ & Density & $k^{*}$ & Density & $\gamma^{*}$ & Density\\
\midrule
Constructive & $9$ & $0.01741$ & $104$ & $0.02101$& $33$ & $0.00920$ & $0.07421$ & $0.02724$\\ 
Cora & $12$ & $0.00842$ & $39$ & $0.00436$ & $74$ & $0.01476$ & $0.02924$ & $0.01242$ \\
AMiner & $8$ & $0.00748$  & $199$ & $0.01786$ & $199$ & $0.03852$ & $0.02317$ & $0.00859$\\
Digits & $5$ & $0.00404$ & $39$ &$0.01400$ & $33$ & $0.01564$ & $0.00346$ & $0.00117$\\
FMA & $1$ & $0.00100$ & $2$ &$0.00103$ & $13$ & $0.00398$ & $0.00146$ & $0.00107$\\
Cell & $1$ & $0.00100$ & $8$ & $0.00133$ & $41$ & $0.00753$ & $0.00320$ & $0.00124$ \\
Segmentation & $7$ & $0.00387$ & $20$ & $0.00637$  & $12$ & $0.00447$ & $0.03423$ & $0.00117$\\
\bottomrule
\end{tabular}}
\end{table}

\begin{figure}[H]
\centering
    \includegraphics[width=0.9\textwidth]{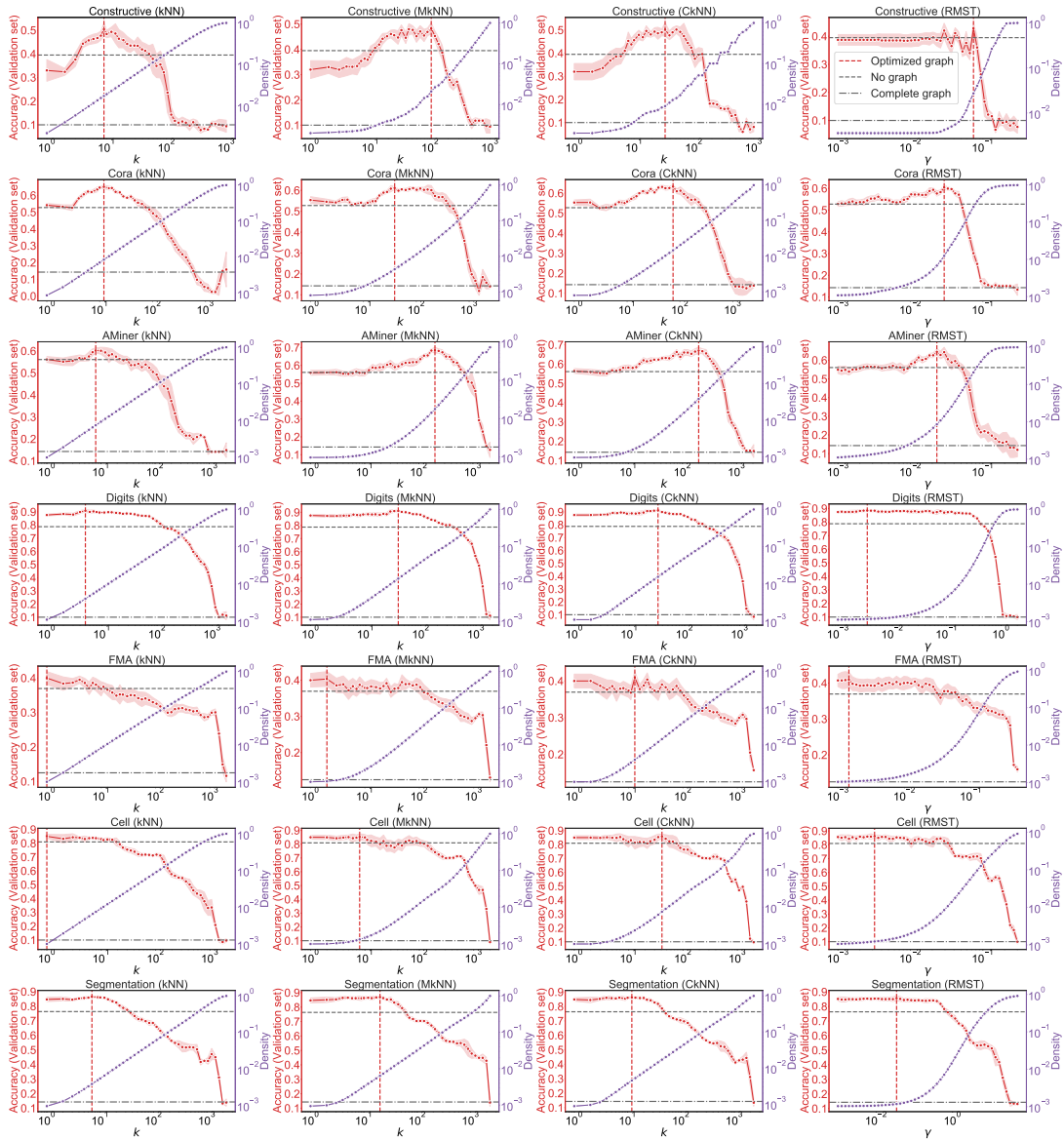}
    \caption{
    Graph construction search in the densification process. The red line indicates the mean classification accuracy on the validation set of $10$ runs with random weight initializations as a function of the density parameter. The red shaded regions denote the standard deviation. The mean classification accuracy on the validation of two limiting cases (no graph and complete graph) are added as well. The red vertical line indicates the optimized graph. The purple line shows the densities of the constructed graphs.}
    \label{fig:densification_density}
\end{figure}

\begin{figure}[H]
\centering
    \includegraphics[width=0.9\textwidth]{allDatasetsDensificationAlignmentMinA1PcaBestCompressed.pdf}
    \caption{The red line indicates the mean classification accuracy on the validation set of $10$ runs with random weight initializations as a function of the density parameter. The red shaded regions denote the standard deviation. The green line indicates the alignment. We report the Pearson correlation coefficients and p-values between mean accuracy and alignment. $\sym{*} \text{p-value}<0.05, \sym{**}\text{p-value}<0.01, \sym{***} \text{p-value}<0.001$.}
    \label{fig:densification_alignment}
\end{figure}

\begin{figure}[H]
\centering
    \includegraphics[width=0.9\textwidth]{allDatasetsDensificationZDistanceCompressed.pdf}
    \caption{The red line indicates the mean classification accuracy on the validation set of $10$ runs with random weight initializations as a function of the density parameter. The red shaded regions denote the standard deviation. The red dashed line represents the mean classification accuracy on the validation of no graph case. The brown line shows the ratio of class separation in the densification process. The brown shaded regions denote the standard deviation. The brown dashed line represents the ratio of class separation of no graph case. We report the Pearson correlation coefficients and p-values between mean accuracy and mean ratio of class separation. $\sym{*} \text{p-value}<0.05, \sym{**}\text{p-value}<0.01, \sym{***} \text{p-value}<0.001$.}
    \label{fig:densification_rcs}
\end{figure}

\begin{figure}[H]
\centering
    \includegraphics[width=0.9\textwidth]{allDatasetsSparsificationValidationDensityCompressed.pdf}
    \caption{
    Graph construction search in the sparsification process. The blue line indicates the mean classification accuracy on the validation set of $10$ runs with random weight initializations as a function of the density parameter. The blue shaded regions denote the standard deviation. The mean classification accuracy on the validation of no graph is added as well. The blue vertical line indicates the optimized graph on the validation set. The purple line shows the densities of the sparsified graphs.
    }
    \label{fig:sparsification_density}
\end{figure}

\begin{table}[H]
\centering
\caption{Comparison between optimized CkNN and sparsification of optimized CkNN graphs (Supplemental Figure~\ref{fig:sparsification_density}).}
\label{table:summary_optimized_graph_sparsification}
\resizebox{0.9\textwidth}{!}{
\begin{tabular}{cccccc|cccc}
\toprule
 \textbf{Top $4$ CkNN graphs} & & \multicolumn{4}{c}{\textbf{Optimized CkNN}} &  \multicolumn{4}{c}{\textbf{Sparsification of optimized CkNN}}\\
 \textbf{on validation set}& \textbf{Data set} & $k^*$  & Edge density& $\langle \text{Degree} \rangle$ & Accuracy (Test) & $\sigma^{*}$& Edge density & $\langle \text{Degree} \rangle$ & Accuracy (Test)\\
\midrule
\multirow{7}{*}{\textbf{(1)}} & Constructive  & $33$ & $0.00920$ & $9.2$ & $51.1$ & $0.3479$ & $0.00630$ & $6.3$ & $51.6$\\
& Cora  & $74$ & $0.01476$ & $36.7$ & $66.6$ & $0$ & $0.01476$ & $36.7$ & $66.6$\\
& AMiner & $199$ & $0.03852$ & $79.8$ & $61.6$ & $0.1618$& $0.01840$ & $38.1$ & $62.5$\\
& Digits & $33$ & $0.01564$ & $28.1$ & $93.4$ & $0$ & $0.01564$ & $28.1$ & $93.4$\\
& FMA  & $13$ & $0.00398$ & $8.0$ & $36.0$ & $0$ & $0.00398$ & $8.0$ & $36.0$\\
& Cell & $41$ & $0.00753$ &  $15.0$ & $84.0$ & $0.5212$  & $0.00240$ & $4.8$ & $85.0$ \\
& Segmentation & $12$ & $0.00447$ & $10.3$  & $83.9$ & $0.3806$ & $0.00356$ & $8.2$ & $84.0$\\
\midrule
& \textbf{Average improvement} & & & & (+8.3) & & & & (+8.7)\\
\midrule
\multirow{7}{*}{\textbf{(2)}} & Constructive  & $51$ & $0.01445$ & $14.4$ & $51.8$ & $0.1898$ & $0.01197$ & $12.0$ & $53.6$\\
& Cora  & $46$ & $0.00897$ & $22.3$ & $66.3$ & $0$ & $0.00897$ & $22.3$ & $66.3$\\
& AMiner & $233$ & $0.04838$ & $100.2$ & $61.3$ & $0.1418$& $0.02396$ & $49.6$ & $63.6$\\
& Digits & $28$ & $0.01319$ & $23.7$ & $93.2$ & $0.4222$ & $0.00556$ & $10.0$ & $93.2$\\
& FMA  & $22$ & $0.00713$ & $14.3$ & $35.2$ & $0.3018$ & $0.00561$ & $11.2$ & $35.8$\\
& Cell & $35$ & $0.00625$ &  $12.5$ & $83.6$ & $0.6409$  & $0.00176$ & $3.5$ & $86.9$ \\
& Segmentation & $7$ & $0.00253$ & $5.8$  & $84.0$ & $0.2607$ & $0.00252$ & $5.8$ & $84.2$\\
\midrule
& \textbf{Average improvement} & & & & (+8.1) & & & & (+9.3)\\
\midrule
\multirow{7}{*}{\textbf{(3)}} & Constructive  & $16$ & $0.00618$ & $6.2$ & $49.0$ & $0$ & $0.00618$ & $6.2$ & $49.0$\\
& Cora  & $39$ & $0.00756$ & $18.8$ & $66.8$ & $0$ & $0.00756$ & $18.8$ & $66.8$\\
& AMiner & 171 &0.03115 &      64.5 &   62.1 &0.1618 & 0.01656 &34.3 &    63.5 \\
& Digits &      21 &0.00978 &      17.6 &   92.9 &0.3425 & 0.00650 &11.7 &    93.0 \\
& FMA  &      41 &0.01457 &      29.1 &   35.9 &     0 & 0.01457 &29.1 &    35.9 \\
& Cell &      48 &0.00904 &      18.1 &   81.9 &0.7806 & 0.00141 & 2.8 &    84.1 \\
& Segmentation &      14 &0.00522 &      12.1 &   83.8 &     0 & 0.00522 &12.1 &    83.8 \\
\midrule
& \textbf{Average improvement} & & & & (+7.7) & & & & (+8.2)\\
\midrule
\multirow{7}{*}{\textbf{(4)}} & Constructive  &      22 &0.00697 &7.0 &   51.2 &0.3084 & 0.00605 & 6.0 &    51.4 \\
& Cora  &      63 &0.01246 &      30.9 &   65.9 &     0 & 0.01246 &30.9 &    65.9 \\
& AMiner &      78 &0.01071 &      22.2 &   62.0 &0.1019 & 0.01021 &21.1 &    62.1 \\
& Digits &      24 &0.01125 &      20.2 &   92.9 &     0 & 0.01125 &20.2 &    92.9 \\
& FMA  &      19 &0.00601 &      12.0 &   34.5 &     0.6808 & 0.00201 &4.0 &    35.2 \\
& Cell &      30 &0.00527 &      10.5 &   81.8 &0.9601 & 0.00099 & 2.0 &    85.3 \\
& Segmentation &      10 &0.00372 &8.6 &   83.9 &0.2607 & 0.00365 & 8.4 &    84.1 \\
\midrule
& \textbf{Average improvement} & & & & (+7.7) & & & & (+8.3)\\
\bottomrule
\end{tabular}
}
\end{table}

\begin{figure}[H]
\centering
    \includegraphics[width=0.9\textwidth]{allDatasetsSparsificationAlignmentMinA1PcaBestCompressed.pdf}
    \caption{The blue line indicates the mean classification accuracy on the validation set of $10$ runs with random weight initializations as a function of the density parameter. The blue shaded regions denote the standard deviation. The green line indicates the alignment. We report the Pearson correlation coefficients and p-values between mean accuracy and alignment. $\sym{*} \text{p-value}<0.05, \sym{**}\text{p-value}<0.01, \sym{***} \text{p-value}<0.001$.}
    \label{fig:sparsification_alignment}
\end{figure}

\begin{figure}[H]
\centering
    \includegraphics[width=0.9\textwidth]{allDatasetsSparsificationZDistanceCompressed.pdf}
    \caption{The blue line indicates the mean classification accuracy on the validation set of $10$ runs with random weight initializations as a function of the density parameter. The blue shaded regions denote the standard deviation. The blue dashed line represents the mean classification accuracy on the validation of no graph case. The brown line shows the ratio of class separation in the sparsification process. The brown shaded regions denote the standard deviation. The brown dashed line represents the ratio of class separation of no graph case. We report the Pearson correlation coefficients and p-values between mean accuracy and mean ratio of class separation. $\sym{*} \text{p-value}<0.05, \sym{**}\text{p-value}<0.01, \sym{***} \text{p-value}<0.001$.}
    \label{fig:sparsification_rcs}
\end{figure}

\begin{table}[H]
    \centering
    \caption{
    Classification accuracy (test set) obtained with three free feature-only methods: MLP, kNN+LDS+GCN~\cite{franceschi2019learning}, and CkNN+GCN (this paper). For information, we also include the accuracy achieved by GCN applied to features together with the additional graph given in the original data set (when available).
    }
    \label{table:results_comparison_LDS}
    \resizebox{1.0\textwidth}{!}{
        \begin{tabular}{c|cccccc|c}
            \toprule
            \textbf{Method}  & \textbf{Cora} & \textbf{AMiner} & \textbf{Digits} & \textbf{FMA} & \textbf{Cell}  & \textbf{Segmentation} & \textbf{Avg. improvement} \\
            \midrule
            MLP &  54.2 &  54.4 & 82.0 & 34.3 & 79.5 & 72.0  & ---\\
            kNN+LDS+GCN~\cite{franceschi2019learning}  &  \textbf{69.0} & 59.3  & \textbf{94.6} & \textbf{36.2}  & 80.0 &  \textbf{83.9}  & (+7.8)\\
            CkNN + GCN (this paper) & 66.6 & \textbf{61.6}  & 93.4 &  36.0 & \textbf{84.0} & \textbf{83.9} & \textbf{(+8.2)}\\
            \midrule
            \textit{Additional original graph + GCN} & \textit{81.1} & \textit{74.8}  & -- &  -- & -- & -- & -- \\
            \bottomrule
    \end{tabular}
    }
\end{table}

\begin{table}[H]
\centering
\caption{
Quality of assignments (test set) obtained by a simple kNN classifier (kNNC), two Louvain-based methods (kNN+Louvain, Seurat),  and our method (CkNN+GCN). The hyperparameters of all methods (kNNC, Louvain methods, and CkNN+GCN) were optimized on the training and validation sets. Two quality measures are computed (ARI and NMI), both normalized between 0 and 1, with higher values indicating better agreement with the ground truth of the test set. The average improvement with respect to the kNNC is also presented in the last column.
}
\label{table:compare_clustering_ARI_NMI}
\resizebox{.9\textwidth}{!}{
    \begin{tabular}{c|cccccc|c}
        \toprule
        \multicolumn{8}{c}{\textbf{ARI}}\\
        \midrule
        \textbf{Method} & \textbf{Cora} & \textbf{AMiner} & \textbf{Digits} & \textbf{FMA} & \textbf{Cell}  & \textbf{Segmentation} & \textbf{Avg. improvement} \\
        \midrule
        kNNC & 0.090 & 0.036 & 0.766 & 0.087 & 0.434 & 0.456 & --- \\
        kNN+Louvain              &  0.337 &   0.301 &   0.840 &  0.086 &  0.721 &         0.273 &  0.115 \\
        Seurat=PCA+kNN+Louvain &  0.321 &   0.305 &   \textbf{0.888} &  0.086 &  \textbf{0.822} &    0.189 &  0.124 \\
        CkNN+GCN (this paper)    &  \textbf{0.382} &   \textbf{0.348} &   0.863 &  \textbf{0.108} &  0.767 &        \textbf{0.702} &  \textbf{0.217} \\
        \bottomrule
        \multicolumn{8}{c}{\textbf{NMI}}\\
        \toprule
        \textbf{Method} & \textbf{Cora} & \textbf{AMiner} & \textbf{Digits} & \textbf{FMA} & \textbf{Cell}  & \textbf{Segmentation} & \textbf{Avg. improvement} \\
        \midrule
        kNNC &0.130& 0.131 & 0.806 & 0.123 & 0.644 & 0.532 & ---\\
        kNN+Louvain              &  0.386 &   0.323 &   0.892 &  0.125 &  0.811 &         0.513 &  0.114 \\
        Seurat=PCA+kNN+Louvain &  0.391 &   0.356 &   \textbf{0.904} &  0.118 &  \textbf{0.904} &         0.350 &  0.110 \\
        CkNN+GCN (this paper)    &  \textbf{0.408} &   \textbf{0.409} &   0.889 &  \textbf{0.147} &  0.854 &         \textbf{0.753} &  \textbf{0.183} \\
        \bottomrule
\end{tabular}
}
\end{table}